\documentclass[journal,comsoc]{IEEEtran}
\usepackage{graphicx}
\usepackage{comment}
\usepackage{amsmath,amssymb} 
\usepackage{color}
\usepackage[normalem]{ulem}

\usepackage{multirow}
\usepackage{booktabs}
\usepackage{url}
\usepackage{footnote}
\usepackage{cite}
\usepackage{subcaption}
\usepackage{soul}

\usepackage{amsthm,amsmath,amssymb}
\usepackage{mathrsfs}
\usepackage{float}
\usepackage{diagbox}

\usepackage[T1]{fontenc}

\usepackage{bm}

\usepackage[capitalise]{cleveref}
\usepackage{soul}

\usepackage[linesnumbered,ruled,vlined]{algorithm2e}

\ifCLASSINFOpdf
\else
\fi
\usepackage{amsmath}
\usepackage[cmintegrals]{newtxmath}
\begin{document}
%
\title{Self-paced Multi-grained Cross-modal Interaction Modeling for Referring Expression Comprehension} 
%
%
%

\author{
  
       Peihan~Miao$^{*}$, Wei~Su$^{*}$, Gaoang Wang, Xuewei~Li$^{\dagger}$, Xi~Li$^{\dagger}$


\IEEEcompsocitemizethanks{\IEEEcompsocthanksitem
Peihan~Miao is with the School of Software Technology, Zhejiang University, Ningbo 315048, China.
Wei~Su, Xuewei~Li, and Xi~Li are with College of Computer Science and Technology, Zhejiang University, Hangzhou 310027, China.
Gaoang Wang is with Zhejiang University-University of Illinois at Urbana-Champaign Institute, Zhejiang University, Haining, 314400, China.
E-mail: \{peihan.miao,~weisuzju,~xueweili,~xilizju\}@zju.edu.cn, gaoangwang@intl.zju.edu.cn.
The first two authors Peihan~Miao and Wei~Su contribute equally.
\protect
}
\thanks{(Correspongding authors: Xuewei~Li and Xi~Li.)}}

\markboth{IEEE TRANSACTIONS ON IMAGE PROCESSING}%
{Shell \MakeLowercase{\textit{et al.}}:Self-paced Multi-grained Cross-modal Interaction Modeling for Referring Expression Comprehension}
%



\maketitle


\begin{abstract}
  As an important and challenging problem in vision-language tasks, referring expression comprehension (REC) generally requires a large amount of multi-grained information of visual and linguistic modalities to realize accurate reasoning.
  In addition, due to the diversity of visual scenes and the variation of linguistic expressions, some hard examples have much more abundant multi-grained information than others. 
  How to aggregate multi-grained information from different modalities and extract abundant knowledge from hard examples is crucial in the REC task.
  To address aforementioned challenges, in this paper, we propose a Self-paced Multi-grained Cross-modal Interaction Modeling framework, which improves the language-to-vision localization ability through innovations in network structure and learning mechanism.
  Concretely, we design a transformer-based multi-grained cross-modal attention, which effectively utilizes the inherent multi-grained information in visual and linguistic encoders.
  Furthermore, considering the large variance of samples, we propose a self-paced sample informativeness learning to adaptively enhance the network learning for samples containing abundant multi-grained information.
  The proposed framework significantly outperforms state-of-the-art methods on widely used datasets, such as RefCOCO, RefCOCO+, RefCOCOg, and ReferItGame datasets, demonstrating the effectiveness of our method.
\end{abstract}

\begin{IEEEkeywords}
Referring expression comprehension, multi-grained cross-modal attention, self-paced sample informativeness learning
\end{IEEEkeywords}

%
\IEEEpeerreviewmaketitle

\vspace{-0.7cm}
  
\setlength{\belowcaptionskip}{-0.2cm}

\section{Introduction}
Referring expression comprehension (REC) \cite{transvg,mattnet,mcn,liu2020attribute,lads,law} is a challenging task that aims at localizing the target object specified by a given referring expression, which potentially underpins a diverse set of vision-language tasks such as referring expression segmentation \cite{mattnet,mcn,liu2021cross,referring-transformer}, visual question answering \cite{antol2015vqa,zhu2016visual7w,chen2023divide}, image captioning \cite{pan2020x,jiang2022visual,huang2021image}, visual commonsense reasoning \cite{zhu2023multi,li2023joint}, and image retrieval \cite{salvador2016faster,yang2023composed,chen2014ranking}.
In addition, the development of REC will contribute to advances in fields such as robotics and human-computer interaction in the physical world \cite{rccf,yang2019fast,rong2019unambiguous,summaira2021recent,jiang2019learning,li2011graph}.

\begin{figure}[htbp]
  \centering
  \includegraphics[height=0.43\textheight,width=0.5\textwidth]{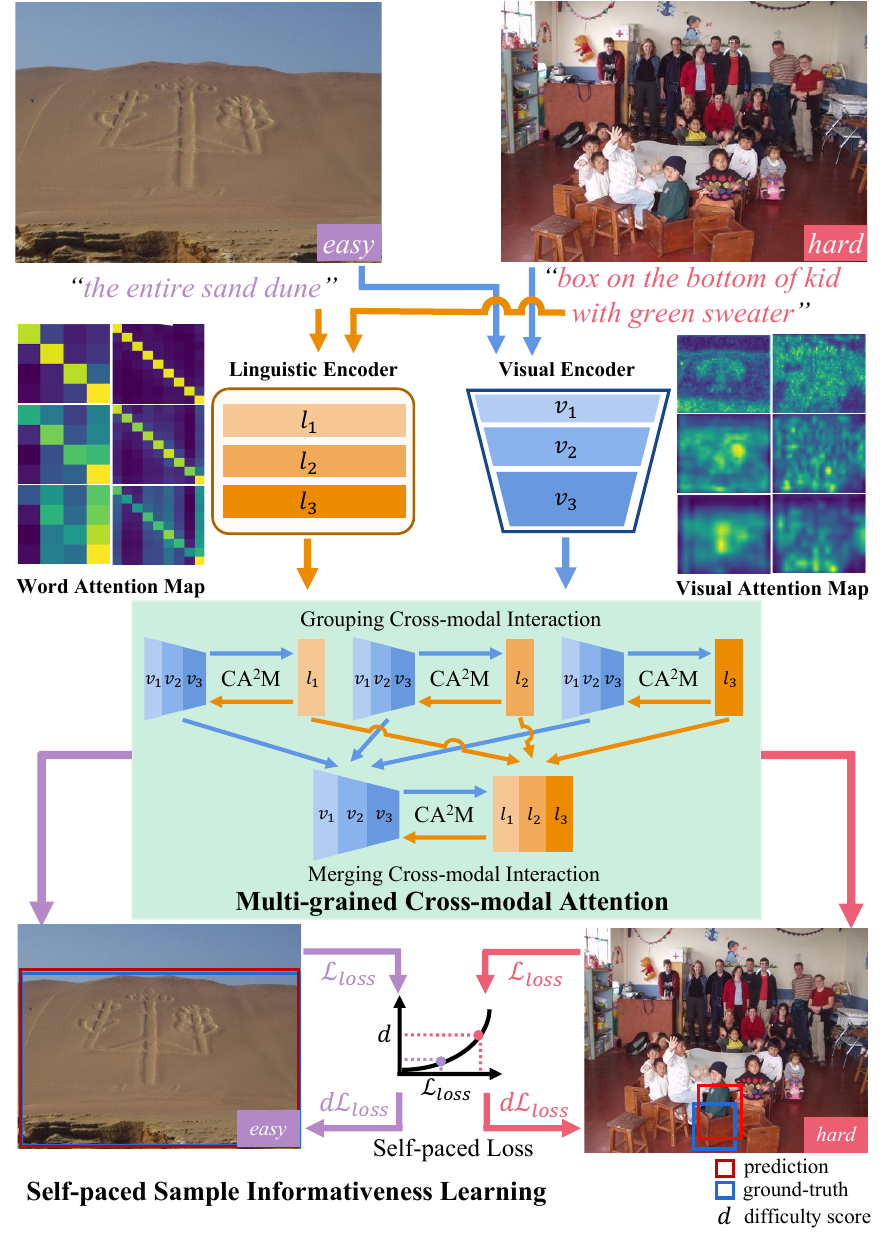}
  \caption{The diagram of the proposed Self-paced Multi-grained Cross-modal Interaction Modeling framework. CA$^{2}$M represents our proposed cross-modal alternate attention module. $v_1$, $v_2$, $v_3$ and $l_1$, $l_2$, $l_3$ represent the visual and linguistic features of the visual and linguistic encoders from shallow to deep layers, respectively.}
  \label{fig:intro}
\end{figure}

As a language-guided detection problem, the REC network generally requires a large amount of multi-grained information of visual and linguistic modalities to realize accurate reasoning.
Specifically, the visual modality needs to obtain visual information including color and texture at the low level, and semantic or relationships of instances at the high level.
The linguistic modality needs to take into account phrase-level information about the descriptive details in phrases, and relatively complex long-range sentence-level information of the whole sentence.
Overlooking multi-grained visual and linguistic information may lead to suboptimal results in the REC, which are verified in our experiments.
A long line of works indicates that visual encoders like ResNet \cite{resnet} tend to capture semantic information in the deeper layers, and detailed information in the shallower layers \cite{yu2017exploiting,liu2016ssd,plv}, as shown in the visual attention maps of \cref{fig:intro}.
Meanwhile, transformer-based \cite{transformer} linguistic encoders like BERT \cite{bert} can capture phrase-level information in the shallower layers, and long-range sentence-level information in the deeper layers \cite{jawahar2019does,plv}, as shown in the word attention maps of \cref{fig:intro}.
These inherent properties of visual and linguistic encoders have great potential to tackle the multi-grained information requirement of REC.
How to aggregate multi-grained information from different modalities and perform multi-grained cross-modal interaction modeling is challenging.
To remedy it, we design a transformer-based multi-grained cross-modal attention, which progressively realizes the interaction modeling between multi-grained visual and linguistic modalities, and achieves accurate and fast language-to-vision localization.

In addition, the learning of samples with abundant multi-grained information can also promote multi-grained cross-modal interaction modeling.
However, the amount of multi-grained information required by the network to realize localization for each sample in REC varies widely.
For instance, in \cref{fig:intro}, compared to the left easy sample that can be easily and intuitively located, the right hard sample, containing complex visual scene and linguistic expression, requires the utilization of multi-grained visual and linguistic information to perform accurate language-to-vision localization.
The diversity of visual scenes and the variation of linguistic expressions make it challenging to intuitively measure the multi-grained informativeness of each sample and customize an adaptive learning strategy for it.
From another perspective, as shown in \cref{fig:intro}, it is difficult for the network to perform multi-grained cross-modal interaction modeling on the samples with abundant multi-grained information, which typically have complex visual scenes and referring expressions.
And the more difficult it is to locate the sample, the more necessary the network needs to utilize multi-grained information to realize localization.
To this end, we use the localization difficulty to reflect the amount of multi-grained information and the necessity of learning for each sample.
Based on the localization difficulty, we design a self-paced sample informativeness learning, which leverages the network to measure the difficulty scores of training samples and adaptively drives the network to acquire knowledge from hard samples with abundant multi-grained information.

\begin{figure}[t]
  \centering
  \begin{subfigure}{0.49\linewidth}
      \centering
      \includegraphics[width=1\linewidth]{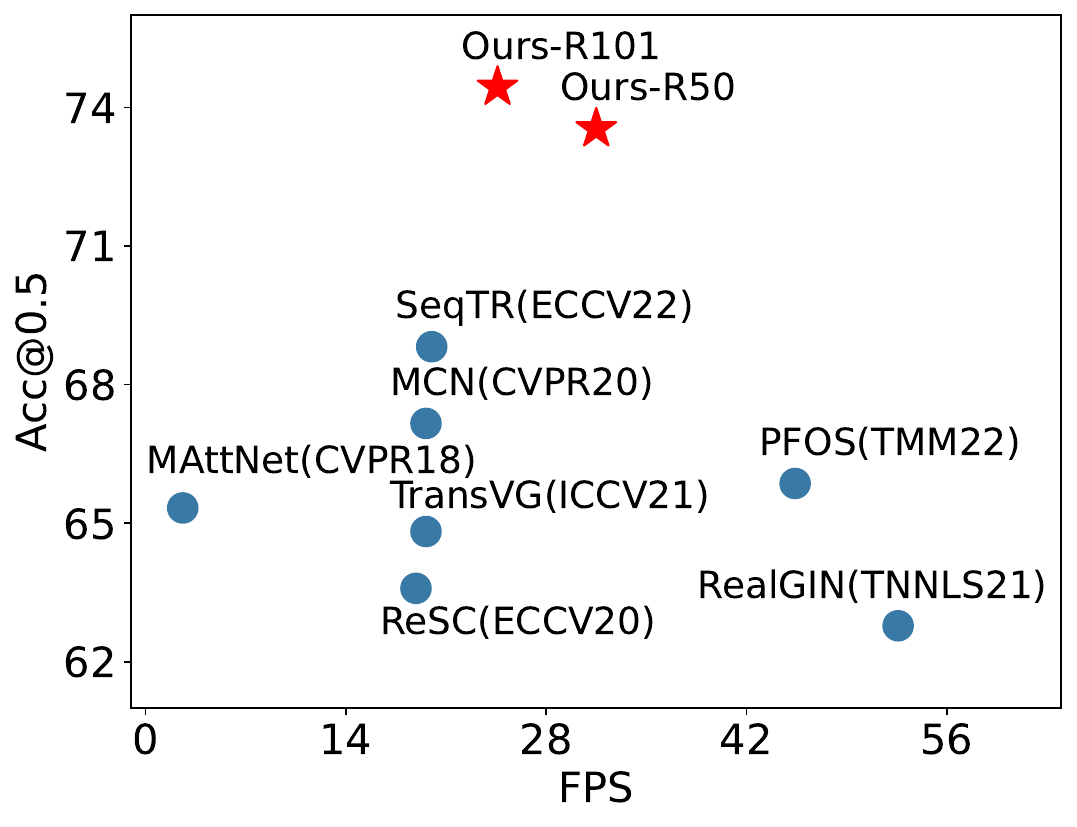}
      \caption{}
  \end{subfigure}
  \begin{subfigure}{0.49\linewidth}
      \centering
      \includegraphics[width=1\linewidth]{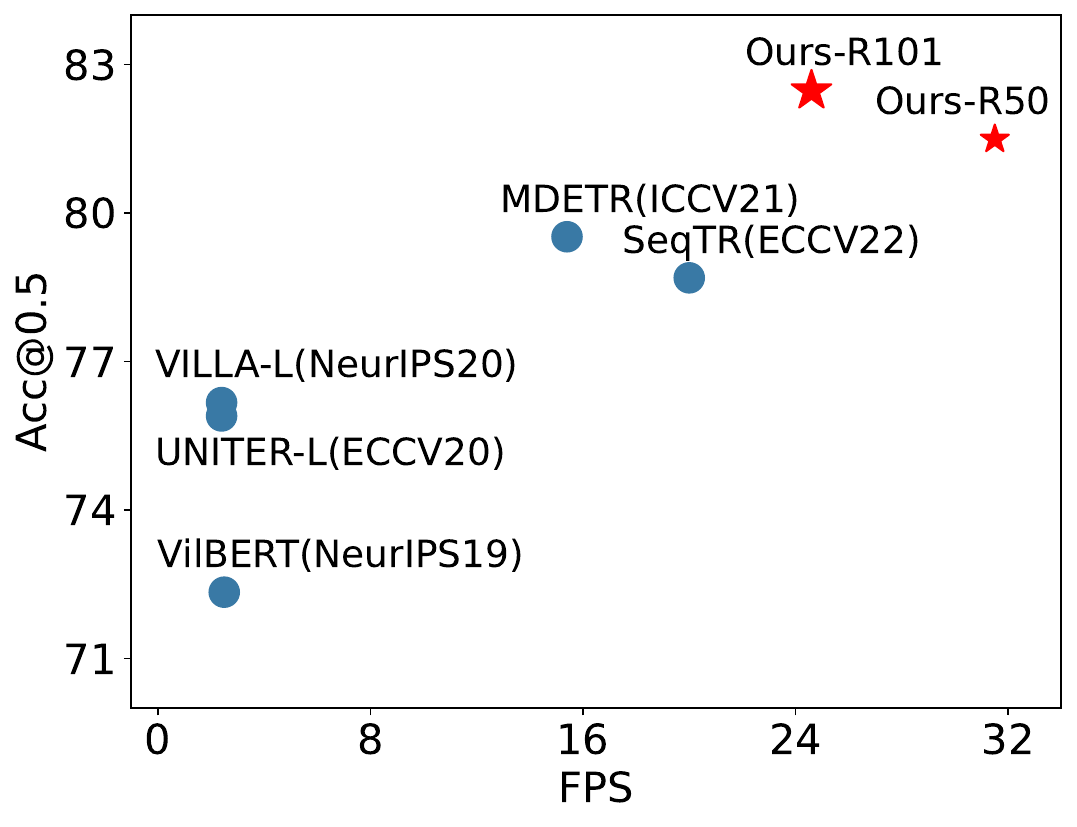}
      \caption{}
  \end{subfigure}

  \centering
  \caption{Comparison of the performance and inference speed on the val set of RefCOCO+.
  (a) and (b) denote comparisons without/with large-scale pre-training, respectively.
  Inference speed is tested on the 1080 Ti.
  }
  \label{fig:fps}
\end{figure}

Based on the aforementioned motivations, we propose a Self-paced Multi-grained Cross-modal Interaction Modeling framework as shown in \cref{fig:intro}.
Concretely, we design a transformer-based multi-grained cross-modal attention, which realizes two processes of grouping and merging cross-modal interactions through our proposed cross-modal alternate attention module (CA$^{2}$M) to utilize the inherent multi-grained information in visual and linguistic encoders.
Furthermore, during each training iteration, we generate difficulty scores for each sample based on the current network.
With the difficulty scores produced by each sample, the network can adaptively learn from samples with abundant multi-grained information. 
Extensive experiments demonstrate the effectiveness of our method, which outperforms state-of-the-art methods by a large margin on the RefCOCO \cite{refcoco}, RefCOCO+ \cite{refcoco}, RefCOCOg \cite{refcocog}, and ReferItGame \cite{referitgame} datasets. 
In addition, with the careful design of the model structure and learning scheme, our method can better balance the performance and inference speed compared with other methods, as shown in \cref{fig:fps}.

Conclusively, our contributions are three-fold:
\begin{itemize}
    \item We propose a Self-paced Multi-grained Cross-modal Interaction Modeling framework, which achieves accurate language-to-vision localization through innovations in network structure and learning mechanism.
    \item We construct a multi-grained cross-modal attention, which effectively utilizes the inherent multi-grained information in visual and linguistic encoders. 
    We design a self-paced sample informativeness learning to adaptively drive network learning on samples with abundant multi-grained information.
    \item Extensive experiments demonstrate the effectiveness of our method, which significantly outperforms state-of-the-art methods on widely-used datasets.
\end{itemize}
\begin{figure*}[htbp]
  \centering
  \includegraphics[height=0.37\textheight,width=\textwidth]{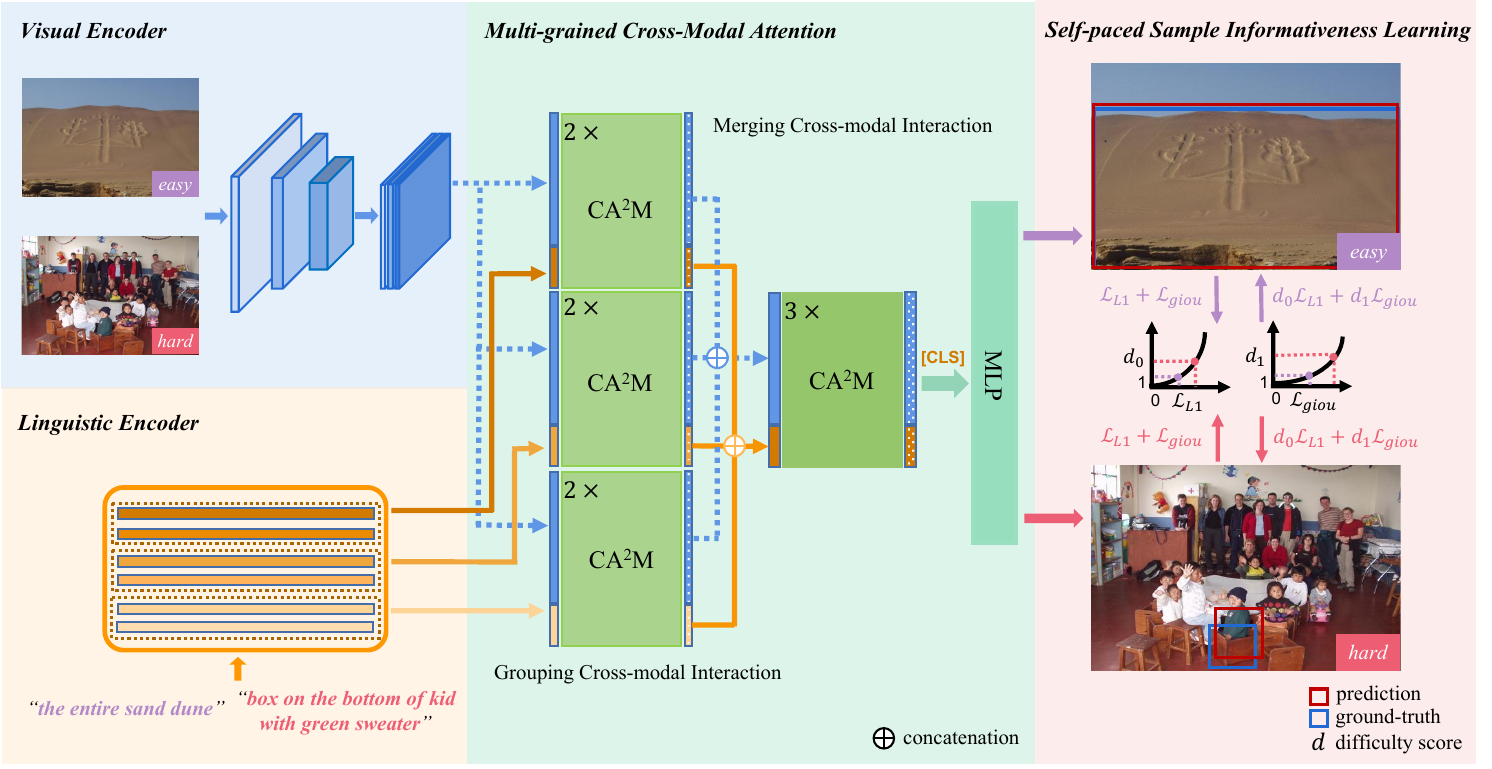}
  \caption{Illustration of the proposed Self-paced Multi-grained Cross-modal Interaction Modeling framework.
  The whole framework consists of the visual encoder, linguistic encoder, multi-grained cross-modal attention, and self-paced sample informativeness learning.
  Concretely, visual and linguistic encoders first extract visual and linguistic features from different layers, respectively.
  In multi-grained cross-modal attention, visual and linguistic features obtained above are used to realize grouping and merging cross-modal interactions through the designed cross-modal alternate attention module (CA$^{2}$M).
  The [CLS] token, derived from linguistic features, aggregates visual and linguistic information during the linguistic feature extraction and multi-grained cross-modal attention, and is finally used for language-to-vision localization.
  A multi-layer perceptron converts the [CLS] token into the coordinates of the target box.
  During the training process, the output coordinates are used to realize self-paced sample informativeness learning.
  Best viewed in color.}
  \label{fig:main}
\end{figure*}

\section{Related Work}
\label{sec2}

\subsection{Referring Expression Comprehension}
\label{related:rec}
Most conventional REC methods include two stages \cite{mattnet,cm-att-erase,ref-nms,liu2019learning,rvg-tree}, \textit{i.e.}, the generation of candidate boxes and the selection of the best matching boxes.
In the first stage, numerous candidate boxes are generated by pre-training object detector, \textit{e.g.} Faster-RCNN \cite{faster}.
In the second stage, the candidate box that best matches the referring expression will be regarded as the target box.
Despite achieving good performance, the two-stage methods are restricted by the accuracy and speed of the candidate boxes, and the matching correctness of the candidate boxes and referring expression.
Recently, state-of-the-art one-stage methods \cite{transvg,word2pix,pfos,plv,referring-transformer,mdetr} have alleviated the limitations of two-stage methods by designing vision-language fusion modules to directly predict the target box.
Despite achieving high performance, such methods typically lack the exploitation of multi-grained information in visual and linguistic encoders.
To this end, we propose a multi-grained cross-modal attention to effectively utilize the inherent multi-grained information in visual and linguistic encoders.

\subsection{Transformer}
\label{related:trans}
Transformer \cite{transformer}, first introduced for neural machine translation, has sparked a huge research boom in recent years.
Transformer-based methods have achieved state-of-the-art performance in natural language tasks \cite{bert,dehghani2018universal,zhu2020incorporating}.
Inspired by the great success in natural language tasks, a large number of transformer-based vision tasks, such as image classification \cite{dosovitskiy2020image,liu2021swin}, object detection \cite{detr,zhu2020deformable} and panoptic segmentation \cite{wang2021max,cheng2021per} have been researched.
In addition, transformer-based cross-modal pre-training methods \cite{gan2020large,chen2020uniter,lu2019vilbert} for vision-language tasks have gradually emerged.
As the core component of the transformer, the attention mechanism \cite{transformer} is developed to capture long-range interactions in the context and the correspondence of tokens across modalities.
Recently, state-of-the-art REC methods \cite{transvg,referring-transformer,mdetr,seqtr,pfos,word2pix} generally use transformer encoder layer \cite{transformer} to capture intra- and inter-modal context across visual and linguistic modalities.
Compared with transformer encoder layer, our proposed CA$^{2}$M can achieve sufficient multi-grained cross-modal interaction while realizing more accurate localization and faster inference.

\subsection{Self-paced Learning}
\label{related:spl}
As a classic progressive learning method \cite{wang2021survey}, self-paced learning \cite{kumar2010self,lin2020pixel,ge2020self,liu2020self,wu2021bispl} uses the current training network to measure the difficulty of the samples and conduct targeted learning.
Different from curriculum learning \cite{bengio2009curriculum,wang2021survey} that requires prior knowledge to determine the sample learning sequences, self-paced learning can dynamically realize sample learning during the training process.
In fact, the samples contained in the REC dataset \cite{refcoco,refcocog,refcocog-umd,referitgame} are diverse, and their visual scenes and referring expressions vary in the amount of multi-grained information.
It requires heavy computational resources and costs to determine the learning strategies manually after enormous trials of experiments.
As a result, we aim to design a self-paced sample informativeness learning to adaptively realize the progressive learning of the network.

\section{Methods}
\label{sec3}
In this section, we illustrate the innovations in the network structure and learning mechanism of the proposed Self-paced Multi-grained Cross-modal Interaction Modeling framework, as shown in \cref{fig:main}.
First, we briefly introduce the overview of motivation and framework in \cref{method:sec1}.
Then, in terms of network structure, we describe the proposed multi-grained cross-modal attention in \cref{method:sec2}, which can effectively utilize multi-grained information with the interaction between visual and linguistic modalities.
Next, in terms of learning mechanism, we measure the multi-grained informativeness of samples in \cref{method:sec3} with difficulty scores, which can form closed-loop feedback in self-paced learning.
Finally, we detail the optimization of the whole framework in \cref{method:sec4}.

\subsection{Overview of Motivation and Framework}
\label{method:sec1}
The REC typically requires a large amount of multi-grained information of visual and linguistic modalities to achieve accurate localization.
Visual and linguistic encoders possess inherent multi-grained information at different layers, and samples with complex visual scenes and referring expressions have abundant multi-grained information.
Inspired by this, we propose a Self-paced Multi-grained Cross-modal Interaction Modeling framework, which improves the reasoning ability through multi-grained cross-modal attention and self-paced sample informativeness learning.

Specifically, given an image and a referring expression, the goal of our framework is to directly regress the coordinates of the target box.
The visual encoder based on ResNet \cite{resnet} and the linguistic encoder based on BERT \cite{bert} first extract multi-grained features from different layers.
In multi-grained cross-modal attention, visual and linguistic features obtained above are used to accomplish grouping and merging cross-modal interactions through the designed cross-modal alternate attention module (CA$^{2}$M).
The [CLS] token, derived from linguistic features, aggregates visual and linguistic information during the linguistic feature extraction and multi-grained cross-modal attention, and is used for localization.
The prediction head composed of a three-layer perceptron converts the [CLS] token into the coordinates of the target box.
During each iteration of training, self-paced sample informativeness learning calculates the difficulty scores of each sample and guides the network to enhance the learning of the samples with abundant multi-grained information.
The framework is shown in \cref{fig:main}.

\subsection{Multi-grained Cross-modal Attention}
\label{method:sec2}
\begin{figure}[t]
    \centering
    \includegraphics[width=0.32\textwidth]{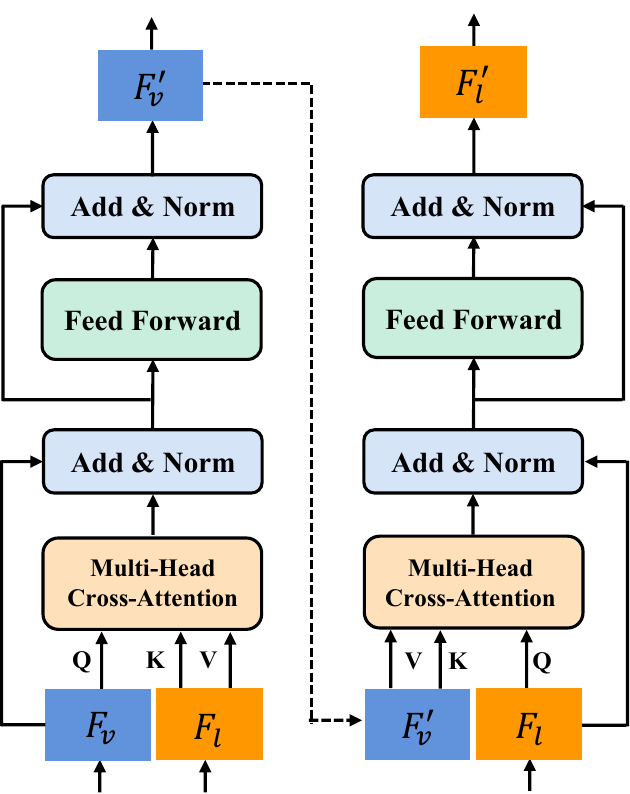}
    \caption{Illustration of the proposed cross-modal alternate attention module (CA$^{2}$M).
    $F_{v}$ and $F_{l}$ represent visual and linguistic features, respectively.}
    \label{fig:ca2m}
\end{figure}

The intrinsic properties of visual and linguistic encoders can provide the multi-grained information required by REC.
Our goal is to aggregate multi-grained information from different modalities and effectively conduct multi-grained cross-modal interaction modeling.
To this end, we design a multi-grained cross-modal attention, which progressively realizes two processes of grouping and merging cross-modal interactions through the proposed cross-modal alternate attention module (CA$^{2}$M), as illustrated in \cref{fig:main} and \cref{fig:ca2m}.

\subsubsection{Cross-modal Alternate Attention Module (CA$^{2}$M)}
The core component of CA$^{2}$M is its attention module, whose input comes from the token sequences of both visual and linguistic modalities.
Specifically, for each head attention module, given the query tokens $X^{q}$ and support tokens $X^{s}$, which are derived from visual and linguistic features or vice versa, respectively, three independent fully-connected layers are used to generate the query embedding $Q$, key embedding $K$ and value embedding $V$. 
We have
\begin{equation}
    \begin{aligned}
    \label{eq1}
    Q=X^{q}W^{q}, \quad K=X^{s}W^{k}, \quad V=X^{s}W^{v}, \\
    \end{aligned}
\end{equation}
where $W^{q}$, $W^{k}$, and $W^{v}$ are the learnable weight matrices for query, key, and value, respectively.
Since $X^{q}$ and $X^{s}$ come from different modalities, the attention module can also be named the cross-attention module.
After obtaining embedding $Q$, $K$, and $V$, the cross-attention module performs cross-attention computation.
We have
\begin{equation}
    \begin{aligned}
    \label{eq2}
    \mathrm{CrossAttention}(Q,K,V)=\mathrm{Softmax}(\frac{QK^{\top}}{\sqrt{d_{\mathrm{dim}}}})V,
    \end{aligned}
\end{equation}
where $d_{\mathrm{dim}}$ is the channel dimension of embeddings.

Visual and linguistic features are fully interacted through the cross-modal attention of CA$^{2}$M, as illustrated in \cref{fig:ca2m}.
Specifically, given visual features $F_{v}$ and linguistic features $F_{l}$, CA$^{2}$M first takes $F_{v}$ and $F_{l}$ as $X^{q}$ and $X^{s}$, respectively, to realize the language-to-vision cross-modal attention, thereby obtaining features $F^{'}_{v}$.
We have
\begin{equation}
    \begin{aligned}
    \label{eq3}
    F^{*}_{v} = \mathrm{LN}(F_{v}+\mathrm{MHCA}(F_{v},F_{l})),
    \end{aligned}
\end{equation}
\begin{equation}
    \begin{aligned}
    \label{eq4}
    F^{'}_{v} = \mathrm{LN}(F^{*}_{v}+\mathrm{FFN}(F^{*}_{v})),
    \end{aligned}
\end{equation}
where $\mathrm{LN}(\cdot)$, $\mathrm{MHCA}(\cdot,\cdot)$ and $\mathrm{FFN}(\cdot)$ represent layer normalization, multi-head cross-attention and feed forward network.
After obtaining $F^{'}_{v}$, CA$^{2}$M regards $F_{l}$ and $F^{'}_{v}$ as $X^{q}$ and $X^{s}$, respectively, thereby obtaining features $F^{'}_{l}$.
We have
\begin{equation}
    \begin{aligned}
    \label{eq5}
    F^{*}_{l} = \mathrm{LN}(F_{l}+\mathrm{MHCA}(F_{l},F^{'}_{v})),
    \end{aligned}
\end{equation}
\begin{equation}
    \begin{aligned}
    \label{eq6}
    F^{'}_{l} = \mathrm{LN}(F^{*}_{l}+\mathrm{FFN}(F^{*}_{l})).
    \end{aligned}
\end{equation}
Through the above cross-modal alternate attention, CA$^{2}$M can achieve sufficient interaction between visual and linguistic modalities.
Compared with transformer encoder layer \cite{transformer} for cross-modal interaction \cite{transvg,mdetr,seqtr,pfos}, CA$^{2}$M can achieve more accurate localization and faster computation, as shown in \cref{tab:cmp}.
Finally, we take CA$^{2}$M as the core component of multi-grained cross-modal attention to realize the following grouping and merging cross-modal interactions.

\begin{figure*}[t]
  \centering
  \includegraphics[width=\textwidth]{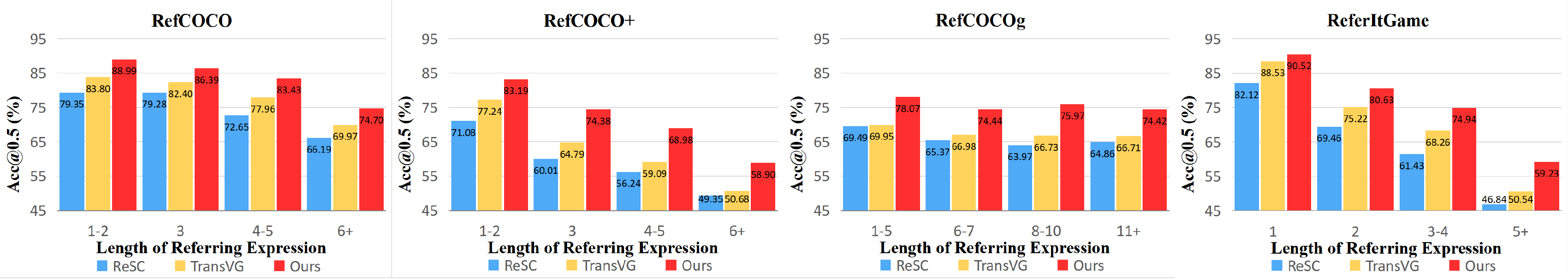}
  \caption{Comparison with state-of-the-art methods \cite{resc,transvg} under different lengths of referring expression on RefCOCO, RefCOCO+, RefCOCOg and ReferItGame test sets.
  }
  \label{fig:length}
\end{figure*}

\subsubsection{Grouping and Merging Cross-modal Interactions}
Multi-grained cross-modal attention is realized by two processes of grouping and merging cross-modal interactions.
To facilitate progressive understanding of visual and linguistic modalities, in grouping cross-modal interaction, visual features containing all multi-grained visual information (\textit{e.g.} low-level and high-level) separately interact with linguistic features containing different multi-grained linguistic information (\textit{e.g.} phrase-level and sentence-level).
In merging cross-modal interaction, all the above-obtained visual and linguistic features are used to perform multi-grained cross-modal interaction and achieve language-guided localization.

To be specific, the visual encoder based on ResNet \cite{resnet} extracts visual features from the last three stages.
Three visual features of different spatial resolutions are transformed to $H \times W \times C_{0}$ by down-sampling or up-sampling, where $C_{0}$, $H$ and $W$ represent the channel, height, and width, respectively. 
Then, visual features are concatenated and flattened to obtain $F^{0}_{v} \in \mathbb{R}^{(HW) \times C_{1}}$.
The linguistic encoder based on six-layer BERT \cite{bert} takes every two layers as a block, and extracts the linguistic features $\{F^{0}_{l,i}\}_{i=1}^{3}$ of the same dimension $L \times C_{1}$, where $L$ and $C_{1}$ represent the length and channel, respectively.
In grouping cross-modal interaction, visual features $F^{0}_{v}$ and linguistic features $F^{0}_{l,i}$ separately interact across modalities through two-layer CA$^{2}$M.
After achieving the first stage of multi-grained cross-modal attention, we obtain $\{F_{v,i}^{1},F_{l,i}^{1}\}_{i=1}^{3}$ with dimension $(HW) \times C_{0}$ and $L \times C_{0}$.
In merging cross-modal interaction, $\{F_{v,i}^{1},F_{l,i}^{1}\}_{i=1}^{3}$ are first concatenated to obtain $F_{v}^{2} \in \mathbb{R}^{(HW) \times C_{1}}$ and $F_{l}^{2} \in \mathbb{R}^{L \times C_{1}}$, respectively.
Three-layer CA$^{2}$M is used to perform the second stage of multi-grained cross-modal attention, resulting in $F_{v}^{3} \in \mathbb{R}^{(HW) \times C_{1}}$ and $F_{l}^{3} \in \mathbb{R}^{L \times C_{1}}$, respectively.
Finally, the [CLS] token obtained from $F_{l}^{3}$ directly predicts the target box $\bm{\hat{b}}=(\hat{x},\hat{y},\hat{w},\hat{h})$ through a three-layer perceptron.

\subsection{Self-paced Sample Informativeness Learning}
\label{method:sec3}
During the training process, extracting abundant knowledge from hard examples can further boost the reasoning ability of the network.
To this end, we design a self-paced sample informativeness learning, which calculates the difficulty scores of training samples and adaptively drives the network to learn samples with abundant multi-grained information.

Concretely, during each iteration of training, we use the classic L1 loss and the generalized IoU loss \cite{giou} as two perspectives \cite{word2pix,vgtr} to measure the difficulty of each sample for the current network.
Given ground-truth $\bm{b}=(x,y,w,h)$ and predicted box $\bm{\hat{b}}=(\hat{x},\hat{y},\hat{w},\hat{h})$, the L1 loss is defined as
\begin{equation}
    \begin{aligned}
    \label{eq7}
    \mathcal{L}_{L1} = |x-\hat{x}| + |y-\hat{y}| + |w-\hat{w}| + |h-\hat{h}|.
    \end{aligned}
\end{equation}
Furthermore, denoting rectangles of $\bm{b}$ and $\bm{\hat{b}}$ as $A$ and $B$, respectively, the generalized IoU loss is defined as
\begin{equation}
    \begin{aligned}
    \label{eq8}
    \mathcal{L}_{giou} = 1 - \frac{|A \cap B|}{|A \cup B|} + \frac{|C-A \cup B|}{|C|},
    \end{aligned}
\end{equation}
where $C$ represents the smallest enclosing rectangle of $A$ and $B$, and $|\cdot|$ represents the area of the region.
By observing equations \cref{eq7} and \cref{eq8}, we can find that the minimum value of $\mathcal{L}_{L1}$ and $\mathcal{L}_{giou}$ is $0$. Simply put, the smaller $\mathcal{L}_{L1}$ and $\mathcal{L}_{giou}$ are, the less difficult the sample is for the current network.

After obtaining $\mathcal{L}_{L1}$ and $\mathcal{L}_{giou}$, we use the function $\mathcal{D}(\cdot)$ to generate difficulty scores $d_{0}$ and $d_{1}$ to guide the network's learning of samples. We have
\begin{equation}
    \begin{aligned}
    \label{eq9}
    d_{0}=\mathcal{D}_{\mu_{0}, \sigma_{0}}(\mathcal{L}_{L1}), \quad  d_{1}=\mathcal{D}_{\mu_{1},\sigma_{1}}(\mathcal{L}_{giou}).
    \end{aligned}
\end{equation}
To be specific, for each loss $\mathcal{L}$, its corresponding difficulty score $d$ is defined with an exponential function as follows:
\begin{equation}
    \begin{aligned}
    \label{eq10}
    d=\mathcal{D}_{\mu, \sigma}(\mathcal{L})=e^{\frac{(\mathcal{L}-\mu)^{2}}{\sigma^{2}}},
    \end{aligned}
\end{equation}
where $\mu$ represents the position of the trough and $\sigma$ controls the width of the function.
Considering that the minimum value of $\mathcal{L}_{L1}$ and $\mathcal{L}_{giou}$ is 0, and the closer $\mathcal{L}_{L1}$ and $\mathcal{L}_{giou}$ are to 0, the more accurate the localization of the sample is, we set both $\mu_{0}$ and $\mu_{1}$ to 0.
In addition, $\sigma_{0}$ and $\sigma_{1}$ are learnable parameters that as $\sigma_{0}$ and $\sigma_{1}$ increase, the functions become smoother.
During each iteration of training, we use $d_{0} \mathcal{L}_{L1}$ and $d_{1} \mathcal{L}_{giou}$ for self-paced sample informativeness learning, which adaptively prompts the network to strengthen the learning of the samples with abundant multi-grained information.

\subsection{Optimization}
\label{method:sec4}
We optimize the proposed Self-paced Multi-grained Cross-modal Interaction Modeling framework end-to-end.
Multi-grained cross-modal attention effectively utilizes the inherent multi-grained information in visual and linguistic encoders.
Self-paced sample informativeness learning leverages localization results to adaptively enhance the learning of the samples with abundant multi-grained information.
The total loss function of our framework is formulated as:
\begin{equation}
    \mathcal{L}_{total} = d_{0} \mathcal{L}_{L1} + d_{1} \mathcal{L}_{giou} - \lambda (d_{0} + d_{1}),
\end{equation}
where $\lambda$ is the regularization coefficient and $\lambda (d_{0} + d_{1})$ is the regularization term to control the $\sigma_{0}$ and $\sigma_{1}$.
The trained framework effectively utilizes multi-grained information to realize more accurate localization, and the experimental analysis of this framework will be elaborated in \cref{sec4}.
\begin{table*}[t]
    \footnotesize
    \centering
    \caption{Comparison with state-of-the-art methods on RefCOCO \cite{refcoco}, RefCOCO+ \cite{refcoco}, RefCOCOg \cite{refcocog} and ReferItGame \cite{referitgame}.
    * represents ImageNet \cite{imagenet} pre-training, and MS-COCO \cite{mscoco} pre-training is used for the rest methods.
    Note that when the visual backbone is pretrained on MS-COCO, overlapping images of the val/test sets of the corresponding datasets are excluded.
    We highlight the best and second best performance in the \textcolor{red}{red} and \textcolor{blue}{blue} colors.
    }
      \begin{tabular}{c|c|c|ccc|ccc|ccc|c}
      \toprule
      \multirow{2}[2]{*}{Methods} & \multirow{2}[2]{*}{Venue} & \multirow{2}[2]{*}{Backbone} & \multicolumn{3}{c|}{RefCOCO} & \multicolumn{3}{c|}{RefCOCO+} & \multicolumn{3}{c|}{RefCOCOg} & ReferItGame \\
            &       &       & val   & testA & testB & val   & testA & testB & val-g & val-u & test-u & test \\
      \midrule
      \textbf{Multi-task:} &       &       &       &       &       &       &       &       &       &       &       &  \\
      MCN \cite{mcn}  & CVPR20 & DarkNet-53/GRU & 80.08 & 82.29 & 74.98 & 67.16 & 72.86 & 57.31 & -     & 66.46 & 66.01 & - \\
      RT* \cite{referring-transformer} & NeurIPS21 & ResNet-101/BERT & 82.23 & 85.59 & 76.57 & 71.58 & 75.96 & 62.16 & -     & 69.41 & 69.40  & 71.42 \\
      PFOS* \cite{pfos} & TMM22 & ResNet-101/BERT & 78.44 & 81.94 & 73.61 & 65.86 & 72.43 & 55.26 & 64.53 & 67.89 & 67.63  & 67.90 \\
      \midrule
      \midrule
      \textbf{Two-stage:} &       &       &       &       &       &       &       &       &       &       &       &  \\
      MAttNet \cite{mattnet} & CVPR18 & ResNet-101/LSTM & 76.65 & 81.14 & 69.99 & 65.33 & 71.62 & 56.02 & -     & 66.58 & 67.27 & 29.04 \\
      RvG-Tree \cite{rvg-tree} & TPAMI19 & ResNet-101/LSTM & 75.06 & 78.61 & 69.85 & 63.51 & 67.45 & 56.66 & -     & 66.95 & 66.51 & - \\
      CM-Att-Erase \cite{cm-att-erase} & CVPR19 & ResNet-101/LSTM & 78.35 & 83.14 & 71.32 & 68.09 & 73.65 & 58.03 & -     & 67.99 & 68.67 & - \\
      Ref-NMS \cite{ref-nms} & AAAI21 & ResNet-101/GRU & 80.70  & 84.00  & 76.04 & 68.25 & 73.68 & 59.42 & -     & 70.55 & 70.62 & - \\
      \midrule
      \midrule
      \textbf{One-stage:} &       &       &       &       &       &       &       &       &       &       &       &  \\
      FAOA \cite{yang2019fast} & ICCV19 & DarkNet-53/BERT & 72.54 & 74.35 & 68.50  & 56.81 & 60.23 & 49.60  & 56.12 & 61.33 & 60.36 & 60.67 \\
      RCCF \cite{rccf}  & CVPR20 & DLA-34/LSTM & -     & 81.06 & 71.85 & -     & 70.35 & 56.32 & -     & -     & 65.73 & 63.79 \\
      ReSC-Large \cite{resc}  & ECCV20 & DarkNet-53/BERT & 77.63 & 80.45 & 72.30  & 63.59 & 68.36 & 56.81 & 63.12 & 67.30  & 67.20  & 64.60 \\
      RealGIN \cite{realgin} & TNNLS21 & DarkNet-53/GRU & 77.25 & 78.70  & 72.10  & 62.78 & 67.17 & 54.21 & -     & 62.75 & 62.33 & - \\
      TransVG \cite{transvg} & ICCV21 & ResNet-101/BERT & 81.02 & 82.72 & 78.35 & 64.82 & 70.70  & 56.94 & 67.02 & 68.67 & 67.73 & 70.73 \\
      VGTR \cite{vgtr} & ICME22 & ResNet-101/LSTM & 79.30 & 82.16 & 74.38 & 64.40  & 70.85  & 55.84 & 64.05 & 66.83 & 67.28 & - \\
      SeqTR \cite{seqtr} & ECCV22 & DarkNet-53/GRU & 81.23 & 85.00 & 76.08 & 68.82  & 75.37  & 58.78 & - & 71.35 & 71.58 & 69.66 \\
      Word2Pix \cite{word2pix} & TNNLS22 & ResNet-101/BERT & 81.20 & 84.39 & 78.12 & 69.74  & 76.11  & 61.24 & - & 70.81 & 71.34 & - \\
      PLV-FPN* \cite{plv} & TIP22 & ResNet-101/BERT & 81.93 & 84.99 & 76.25 & 71.20  & 77.40  & 61.08 & - & 70.45 & 71.08 & 71.77 \\
      \textbf{Ours}  & -     & ResNet-50/BERT & \textcolor{blue}{84.71} & \textcolor{blue}{87.91} & \textcolor{blue}{79.49} & \textcolor{blue}{73.54} & \textcolor{blue}{79.26}  & \textcolor{blue}{63.60} & \textcolor{blue}{73.87}  & \textcolor{blue}{75.61} & \textcolor{blue}{74.91} & \textcolor{blue}{74.55} \\
      \textbf{Ours}  & -     & ResNet-101/BERT & \textcolor{red}{85.10} & \textcolor{red}{88.23} & \textcolor{red}{80.08} & \textcolor{red}{74.44} & \textcolor{red}{79.48} & \textcolor{red}{65.21} & \textcolor{red}{74.50} & \textcolor{red}{77.25} & \textcolor{red}{75.78} & \textcolor{red}{75.18} \\
      \bottomrule
      \end{tabular}%
    \label{tab:sota}%
  \end{table*}%
\begin{table*}[t]
    \centering
    \caption{Comparison with state-of-the-art large-scale pre-training methods on RefCOCO \mbox{\cite{refcoco}}, RefCOCO+ \mbox{\cite{refcoco}} and RefCOCOg \mbox{\cite{refcocog}}.
    We highlight the best and second best performance in the \textcolor{red}{red} and \textcolor{blue}{blue} colors.}
      \begin{tabular}{c|c|c|c|ccc|ccc|cc}
      \toprule
      \multirow{2}[2]{*}{Models} & \multirow{2}[2]{*}{Venue} & \multirow{2}[2]{*}{Backbone} & \multirow{2}[2]{*}{Pre-train Images} & \multicolumn{3}{c|}{RefCOCO} & \multicolumn{3}{c|}{RefCOCO+} & \multicolumn{2}{c}{RefCOCOg} \\
            &       &       &       & val   & testA & testB & val   & testA & testB & val-u & test-u \\
      \midrule
      VilBERT \cite{lu2019vilbert} & NeurIPS19 & ResNet-101/BERT & 3.3M  &    -   &   -    &   -    & 72.34 & 78.52 & 62.61 &    -   & - \\
      UNITER-L \cite{chen2020uniter} & ECCV20 & ResNet-101/BERT & 4.6M  & 81.41 & 87.04 & 74.17 & 75.90  & 81.45 & 66.70  & 74.86 & 75.77 \\
      VILLA-L \cite{gan2020large} & NeurIPS20 & ResNet-101/BERT & 4.6M  & 82.39 & 87.48 & 74.84 & 76.17 & 81.54 & 66.84 & 76.18 & 76.71 \\
      ERNIE-ViL-L \cite{yu2021ernie} & AAAI21 & ResNet-101/BERT & 4.3M  &   -    &   -    &   -    & 75.89 & 82.37 & 66.91 &   -    & - \\
      MDETR \cite{mdetr} & ICCV21 & ResNet-101/RoBERTa & 200K  & 86.75 & 89.58 & 81.41 & 79.52 & 84.09 & 70.62 & 81.64 & 80.89 \\
      SeqTR \cite{seqtr} & ECCV22 & DarkNet-53/GRU & 174K  & 87.00 & 90.15 & 83.59 & 78.69 & 84.51 & 71.87 & 82.69 & 83.37 \\
      \textbf{Ours}  &   -   & ResNet-50/BERT & 174K  &   \textcolor{blue}{89.15}    &   \textcolor{blue}{91.64}    &   \textcolor{blue}{86.18}    &   \textcolor{blue}{81.49}    &   \textcolor{blue}{86.36}    &   \textcolor{blue}{75.31}    &   \textcolor{blue}{84.23}    & \textcolor{blue}{84.46} \\
      \textbf{Ours}  &   -   & ResNet-101/BERT & 174K  &   \textcolor{red}{89.60}    &   \textcolor{red}{92.26}    &   \textcolor{red}{86.69}    &   \textcolor{red}{82.47}    &   \textcolor{red}{87.32}    &  \textcolor{red}{75.54}     &   \textcolor{red}{85.40}    & \textcolor{red}{85.25} \\
      \bottomrule
      \end{tabular}%
    \label{tab:sota_pretrain}%
  \end{table*}%

\section{Experiments}
\label{sec4}

\subsection{Datasets and Evaluation Protocol}

\subsubsection*{Datasets}
For the task of REC, we conduct experiments with the widely used RefCOCO \cite{refcoco}, RefCOCO+ \cite{refcoco}, RefCOCOg \cite{refcocog}, and ReferItGame \cite{referitgame} datasets.
RefCOCO, RefCOCO+, and RefCOCOg are collected from MS-COCO \cite{mscoco}, whereas ReferItGame is collected from SAIAPR-12 \cite{saiapr12}.
RefCOCO and RefCOCO+ are obtained from the interactive game interface, including the train, val, testA, and testB sets.
RefCOCOg is collected in a non-interactive setting, which has two split datasets, RefCOCOg-google \cite{refcocog} and RefCOCOg-umd \cite{refcocog-umd}, where RefCOCOg-google contains the train and val-g sets, and RefCOCOg-umd has the train, val-u and test-u sets.
RefCOCOg is generated in a non-interactive setting, providing longer referring expressions than RefCOCO and RefCOCO+.
Following a cleaned version of split \cite{rohrbach2016grounding}, ReferItGame has the train and test sets.
We use the val set of RefCOCOg-google for ablation analysis, and use other sets for performance comparisons with state-of-the-art methods.

Furthermore, state-of-the-art methods \cite{mdetr,seqtr,yu2021ernie,gan2020large,chen2020uniter,lu2019vilbert} achieve superior performance using large-scale pre-training.
To demonstrate the effectiveness of our method, we also use the large-scale dataset \cite{seqtr} for pre-training, where the train sets of VG \cite{vg}, RefCOCO \cite{refcoco}, RefCOCO+ \cite{refcoco} and RefCOCOg \cite{refcocog,refcocog-umd}, ReferItGame \cite{referitgame} and Flickr \cite{flickr30k} datasets are included.
After large-scale pre-training, we separately use the train set of RefCOCO \cite{refcoco}, RefCOCO+ \cite{refcoco} and RefCOCOg \cite{refcocog} datasets for fine-tuning.

\subsubsection*{Evaluation Protocol}
Following the previous works \cite{transvg,rccf,resc}, we use \emph{Acc}@0.5 to evaluate the performance of the network.
Concretely, given an image and a referring expression, a predicted bounding box is regarded as correct if the intersection-over-union (IoU) with the ground-truth bounding box is greater than 0.5.
In addition, we also report the inference speed of our proposed method.

\subsection{Implementation Details}
\subsubsection*{Training}
The whole network is end-to-end optimized by AdamW \cite{adamw} with a weight decay of 1e-4.
ResNet-50 and ResNet-101 \cite{resnet} are used as the visual encoders to extract visual features from the last three stages.
The visual encoders used in the experiment are provided by TranVG \cite{transvg}, which uses DETR \cite{detr} for pre-training on the MS-COCO dataset.
In particular, during pre-training of the visual encoder, images in the val/test sets of RefCOCO/+/g overlapping with MS-COCO are removed.
The uncased base of six-layer BERT \cite{bert} pre-trained on the BookCorpus \cite{zhu2015aligning} and English Wikipedia \cite{bert} datasets is used as a linguistic encoder to extract linguistic features.
$\sigma_{0}$ and $\sigma_{1}$ are initialized to 2.5, and $\lambda$ is set to 0.75.
$H$ and $W$ are 16 times smaller than the input image, and $C_{0}$ and $C_{1}$ are set to 256 and 768, respectively.
For the CA$^2$M in grouping and merging cross-modal interactions, the dimensions of the multi-head cross attention are 256 and 768, the numbers of attention head are 8 and 12, the intermediate sizes of the feedforward layer are 1024 and 3072, respectively.
We set an initial learning rate of 1e-5 for the visual and linguistic encoders, 5e-5 for $\sigma_{0}$ and $\sigma_{1}$, and 1e-4 for the remaining components.
The model is trained for 120 epochs with a batch size of 192, where the learning rate is reduced by a factor of 10 at 80 epochs.
In large-scale pre-training and fine-tuning, we train for 40 and 60 epochs, where the learning rate is reduced by 10 at 30 and 40 epochs, respectively.
The input image is resized to $512 \times 512$. Data augmentation operations include random horizontal flips, random affine transformations, random color space jittering (saturation and intensity), and Gaussian blur.
The maximum length of referring expression is set to 40, where the [CLS] and [SEP] tokens are inserted into the referring expression.
We implement our framework with PyTorch and conduct experiments with eight NVIDIA A100 GPUs.
\subsubsection*{Inference}
The input image is resized to $512 \times 512$ and the maximum length of referring expressions is set to 40.
There is no data augmentation during inference.
Our framework directly outputs the coordinates of the target box without any post-processing operations.

\begin{table}[t]
    \footnotesize
    \centering
    \caption{Ablation study of the number of visual and linguistic features in our framework on the RefCOCOg-google val set.}
      \begin{tabular}{ccc}
      \toprule
      Number of visual features & Number of linguistic features & \emph{Acc}@0.5 \\
      \midrule
      \small{One layer}     & \small{One layer}     & \small{66.52} \\
      \small{Three layers}     & \small{One layer}     & \small{70.08} \\
      \small{One layer}     & \small{Three layers}     & \small{70.16} \\
      \small{Three layers}      & \small{Two layers}     & \small{72.48} \\
      \small{Two layers}      &  \small{Three layers}      & \small{72.61} \\
      \small{Three layers}     &  \small{Three layers}      & \small{\textbf{72.86}} \\
      \bottomrule
      \end{tabular}%
    \label{tab:number_features}%
  \end{table}%
  
\begin{table}[t]
    \small
    \centering
    \caption{Ablation study of the number of CA$^{2}$M in our framework on the RefCOCOg-google val set.}
      \begin{tabular}{ccc}
      \toprule
      \normalsize{Number of grouping} & \normalsize{Number of merging} & \normalsize{\emph{Acc}@0.5} \\
      \midrule
      Three layers     & One layer     & 69.60 \\
      Three layers     & Two layers     & 71.89 \\
      Six layers     & One layer     &  71.78\\
      Six layers      & Two layers     & 72.59 \\
      Zero layer      & Nine layers     & 71.72 \\
      Six layers     &  Three layers      & \textbf{72.86} \\
      \bottomrule
      \end{tabular}%
    \label{tab:number_layers}%
  \end{table}%

\begin{figure}[t]
  \centering
  \begin{subfigure}{0.49\linewidth}
      \centering
      \includegraphics[width=1\linewidth]{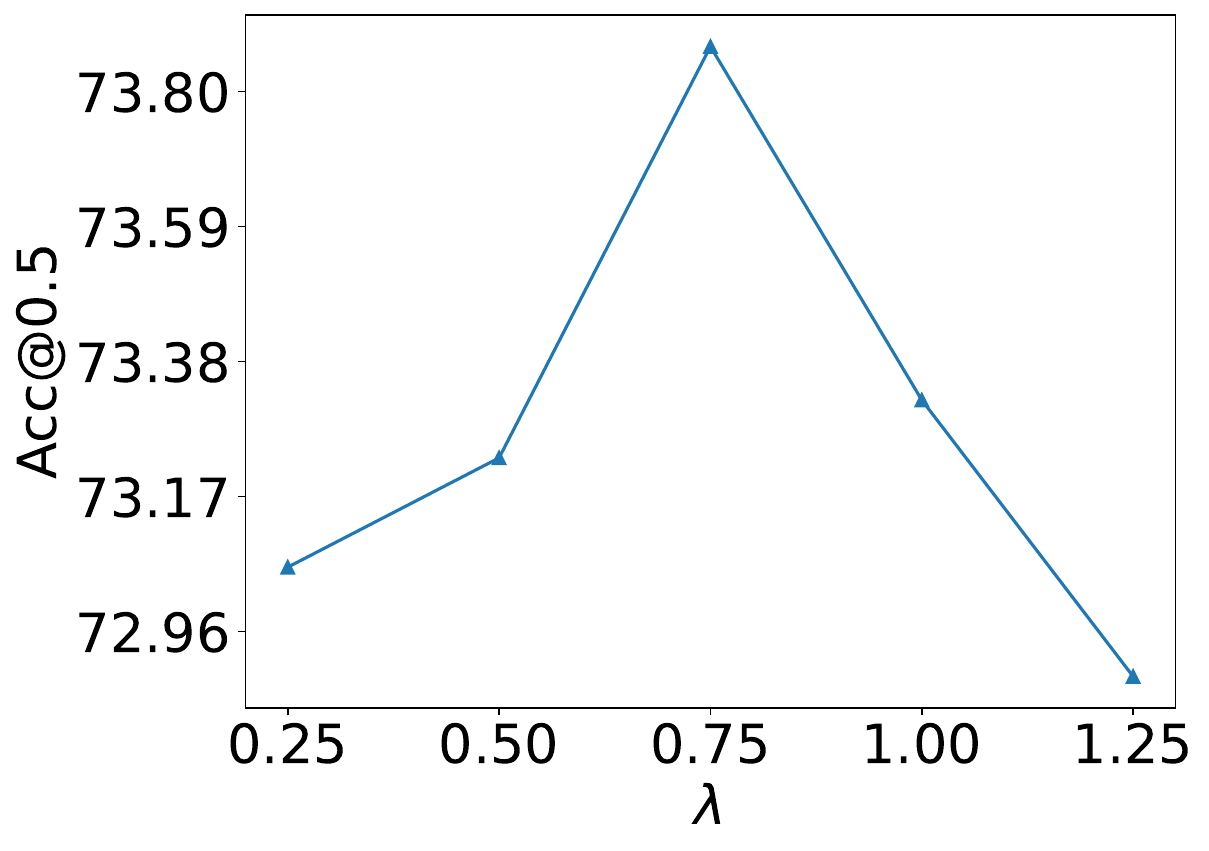}
  \end{subfigure}
  \begin{subfigure}{0.49\linewidth}
      \centering
      \includegraphics[width=1\linewidth]{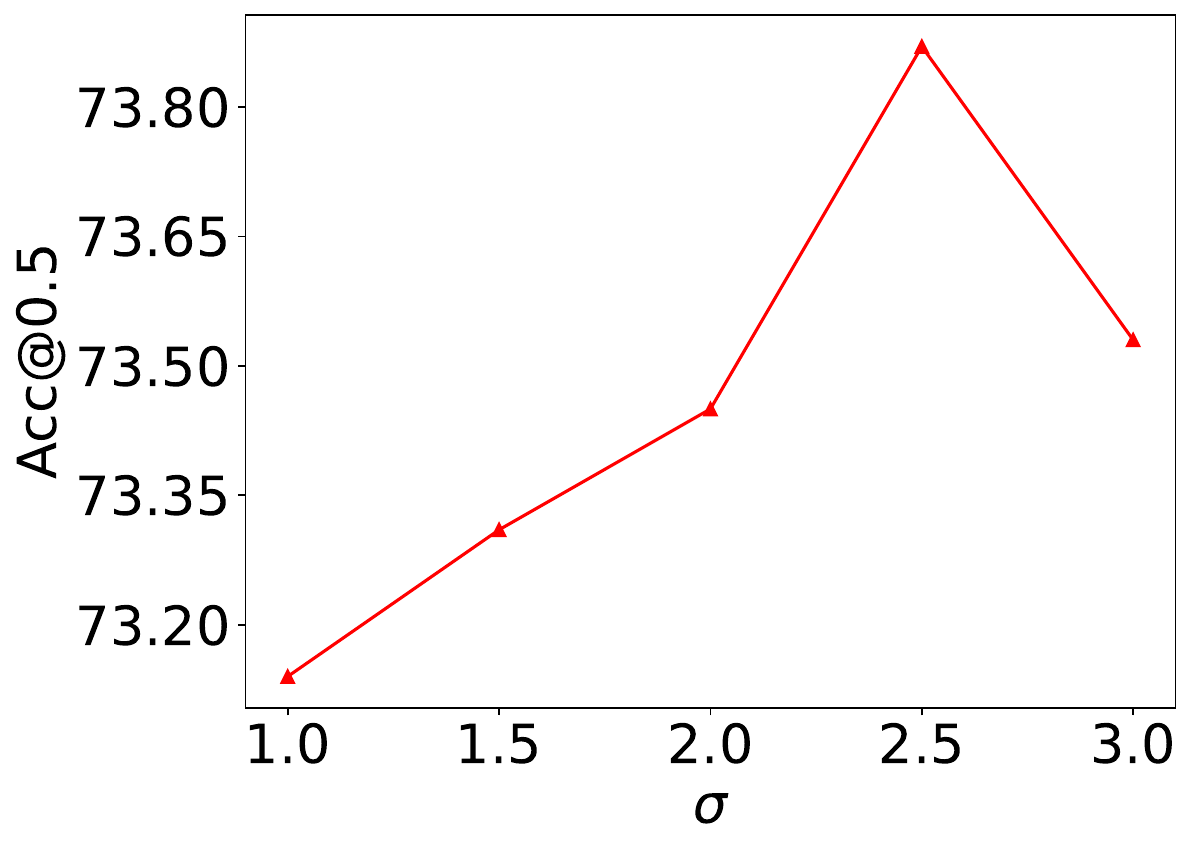}
  \end{subfigure}

  \centering
  \caption{Ablation study of $\lambda$ and $\sigma$ in self-paced sample informativeness learning on the RefCOCOg-google val set.
  }
  \label{fig:lambda_sigma}
\end{figure}

\captionsetup[figure]{name={Fig.},labelsep=period,singlelinecheck=off} 

\begin{figure}[t]
  \centering
  \begin{subfigure}{0.49\linewidth}
      \centering
      \includegraphics[width=1\linewidth]{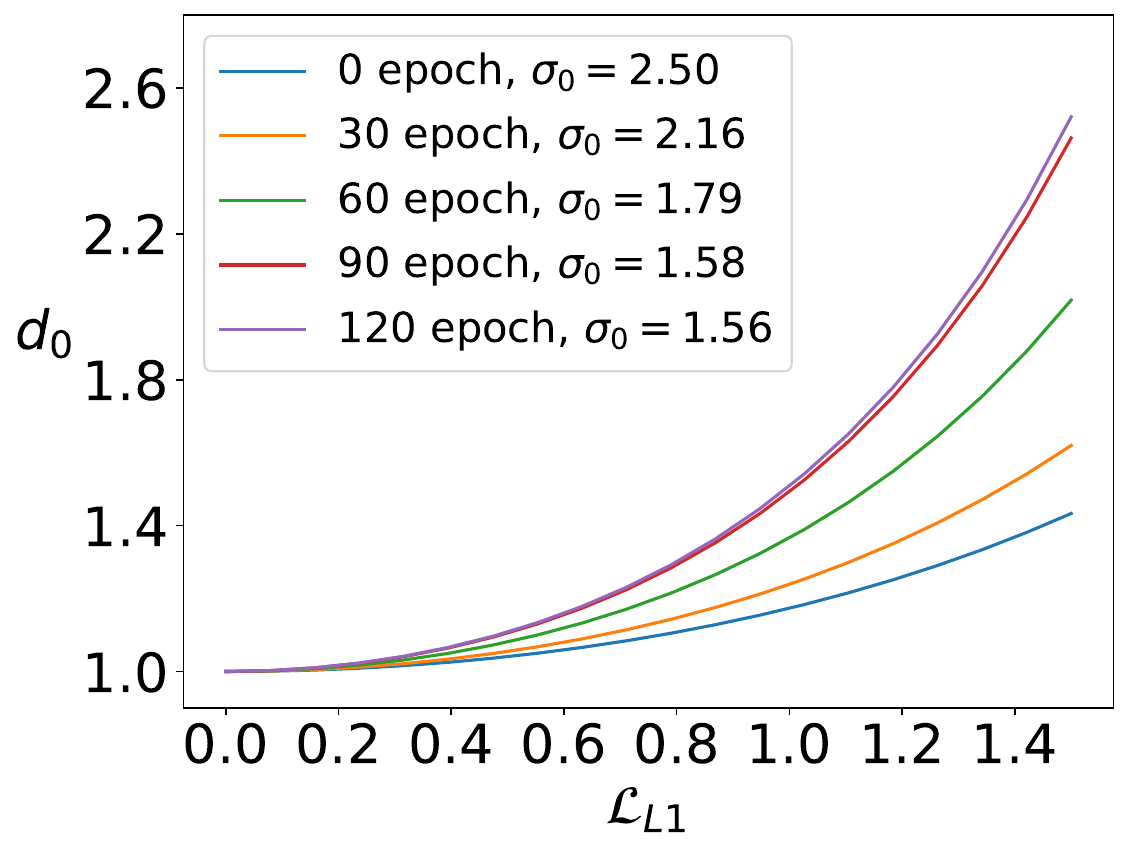}
  \end{subfigure}
  \begin{subfigure}{0.49\linewidth}
      \centering
      \includegraphics[width=1\linewidth]{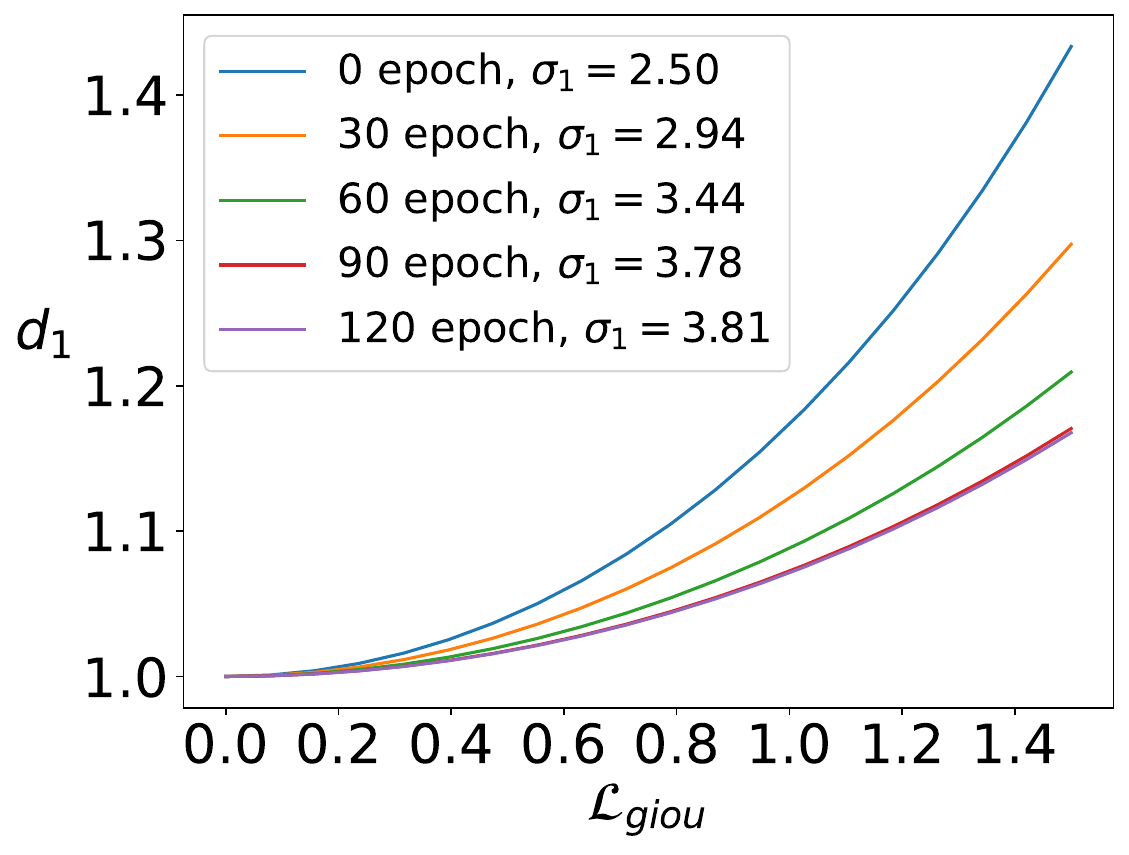}
  \end{subfigure}

  \centering
  \caption{The variation of the $e^{\frac{(\mathcal{L}-\mu)^{2}}{\sigma^{2}}}$ during training.
  }
  \label{fig:epoch_sigma}
\end{figure}

\begin{figure*}[t]
  \centering
  \includegraphics[height=0.27\textheight,width=\textwidth]{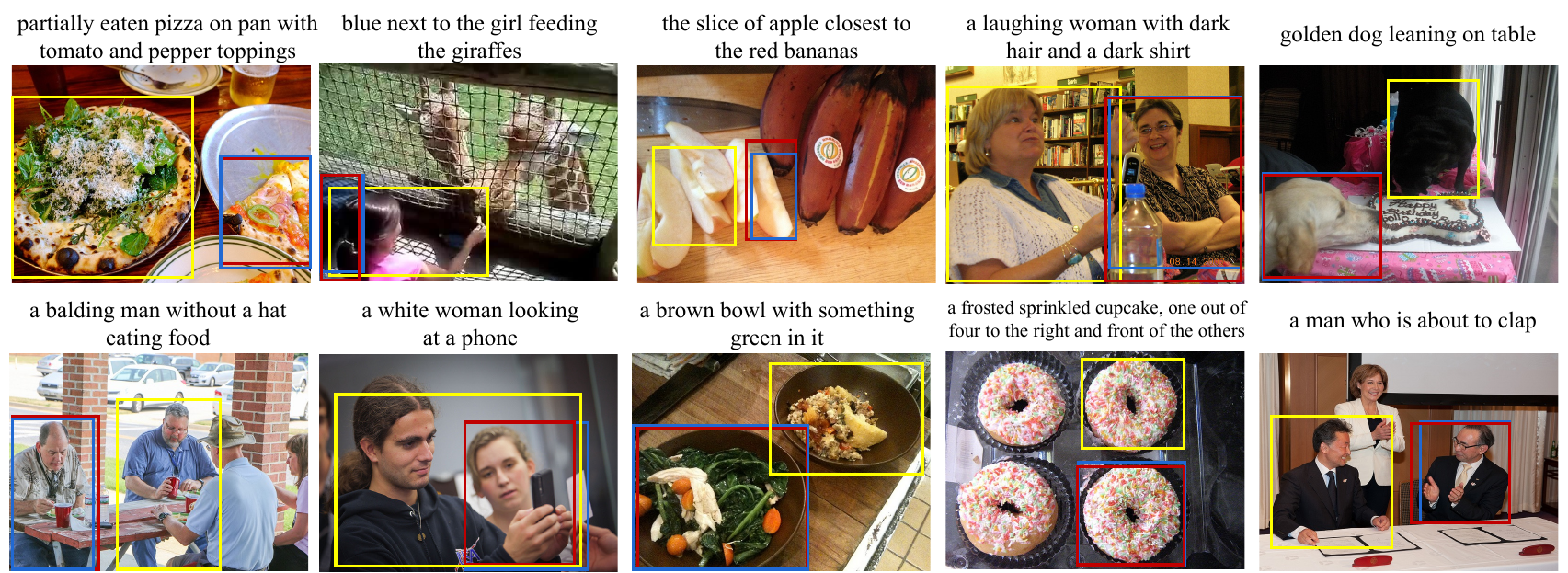}
  \caption{Qualitative results of the proposed framework on the RefCOCOg-google val set.
  The \textcolor{red}{red} boxes represent the predictions of our framework, the \textcolor{yellow}{yellow} boxes represents the predictions of our baseline, and the \textcolor{blue}{blue} boxes are the corresponding ground-truths.
  Best viewed in color.}
  \label{fig:visual}
\end{figure*}

\begin{table*}[h]
  \scriptsize
  \centering
  \caption{Comparison of CA$^{2}$M and transfomer encoder layer on RefCOCO, RefCOCO+, RefCOCOg and ReferItGame.}
    \begin{tabular}{c|ccc|ccc|ccc|c|ccc}
    \toprule
    Cross-modal  & \multicolumn{3}{c|}{RefCOCO} & \multicolumn{3}{c|}{RefCOCO+} &       & RefCOCOg &       & ReferItGame & \multicolumn{3}{c}{Model} \\
    Interaction Block & val   & testA & testB & val   & testA & testB & val-g & val-u & test-u & test  & FPS$_{\uparrow}$   & FLOPS$_{\downarrow}$ & Params$_{\downarrow}$ \\
    \midrule
    Transfomer Encoder &   82.68    &   85.81    &   77.35    &   71.84    &   77.60    &   62.38    &  70.83     &   72.04    &    72.30   &   71.43    & 25.3  & 126.89G & \textbf{121.14M} \\
    CA$^{2}$M  & \textbf{84.71} & \textbf{87.91} & \textbf{79.49} & \textbf{73.54} & \textbf{79.26} & \textbf{63.60}  & \textbf{73.87} & \textbf{75.61} & \textbf{74.91} & \textbf{74.55} & \textbf{31.5}  & \textbf{110.76G} & 147.15M \\
    \bottomrule
    \end{tabular}%
  \label{tab:cmp}%
\end{table*}%

\subsection{Comparisons with State-of-the-art Methods}
To verify the effectiveness of our proposed framework, we first report performance comparisons without/with large-scale pre-training.
Then, we compare the performance with state-of-the-art methods under different lengths of referring expression.
Finally, the performance and inference speed of our method are also compared with other REC methods.

\cref{tab:sota} shows the comparison of our framework with state-of-the-art one-stage methods \cite{plv,word2pix,seqtr,vgtr,transvg,realgin,resc,rccf,yang2019fast}.
Compared to TransVG \cite{transvg}, we observe a substantial performance boost of +4.08\%/ +5.51\%/ +1.73\% on RefCOCO val, testA, and testB sets, +9.62\%/ +8.78\%/ +8.27\% on RefCOCO+ val, testA, and testB sets, +7.48\%/ +8.58\%/ +8.05\% on RefCOCOg val-g, val-u and test-u sets, and +4.45\% on ReferItGame test set.
A comparison of our framework with state-of-the-art two-stage methods \cite{ref-nms,cm-att-erase,rvg-tree,mattnet} is shown in \cref{tab:sota}.
Compared to Ref-NMS \cite{ref-nms}, we observe the performance improvement of +4.40\%/ +4.23\%/ +4.04\% on RefCOCO val, testA, and testB sets, +6.19\%/ +5.80\%/ +5.79\% on RefCOCO+ val, testA, and testB sets, and +6.70\%/ +5.16\% on RefCOCOg val-u and test-u sets.
Besides, \cref{tab:sota} also shows the comparison of our method with state-of-the-art multi-task learning methods \cite{mcn,referring-transformer,pfos}.
MCN \cite{mcn} and RT \cite{referring-transformer} are multi-task learning methods including REC and referring expression segmentation (RES) while PFOS \cite{pfos} is a multi-task learning method including REC and referring expression generation (REG).
Compared to TR \cite{referring-transformer}, we observe the performance improvement of +2.87\%/ +2.64\%/ +3.51\% on RefCOCO val, testA, and testB sets, +2.86\%/ +3.52\%/ +3.05\% on RefCOCO+ val, testA, and testB sets, +7.84\%/ +6.38\% on RefCOCOg val-u and test-u sets, and +3.76\% on ReferItGame test set.

We also compare the performance with state-of-the-art large-scale pre-training methods \cite{seqtr,mdetr,yu2021ernie,gan2020large,chen2020uniter,lu2019vilbert} on the RefCOCO \cite{refcoco}, RefCOCO+ \cite{refcoco} and RefCOCOg \cite{refcocog} datasets, as shown in \cref{tab:sota_pretrain}.
Compared to SeqTR \cite{seqtr}, we observe the performance improvement of +2.60\%/ +2.11\%/ +3.10\% on RefCOCO val, testA, and testB sets, +3.78\%/ +2.81\%/ +3.67\% on RefCOCO+ val, testA, and testB sets, and +2.71\%/ +1.88\% on RefCOCOg val-u and test-u sets.

\cref{fig:length} shows the performance comparison of state-of-the-art methods \cite{transvg,resc} under different lengths of referring expression, where TransVG \cite{transvg} and ReSC \cite{resc} with open-source models are selected for comparison.
Compared to TransVG \cite{transvg}, we observe a substantial performance boost of +5.91\%/ +3.99\%/ +5.47\%/ +4.73\% on RefCOCO test set, +5.95\%/ +9.59\%/ +9.89\%/ +8.22\% on RefCOCO+ test set, +8.12\%/ +7.46\%/ +9.24\%/ +7.71\% on RefCOCOg test set and +1.99\%/ +5.41\%/ +6.68\%/ +8.69\% on ReferItGame test set from short to long referring expressions.

Furthermore, we compare the performance and inference speed of the proposed method with other REC methods on the val set of RefCOCO+, as shown in \cref{fig:fps}.
Inference speed is tested on the 1080 Ti, and our method achieves 31.5 and 24.6 FPS based on ResNet-50 and ResNet-101, respectively.
\cref{fig:fps} (a) and (b) show the comparison without/with large-scale pre-training, respectively.
We can find that our proposed method achieves excellent performance as well as fast inference speed.

\begin{figure*}[t]
  \centering
  \captionsetup{justification=centering}
  \includegraphics[height=0.36\textheight,width=\textwidth]{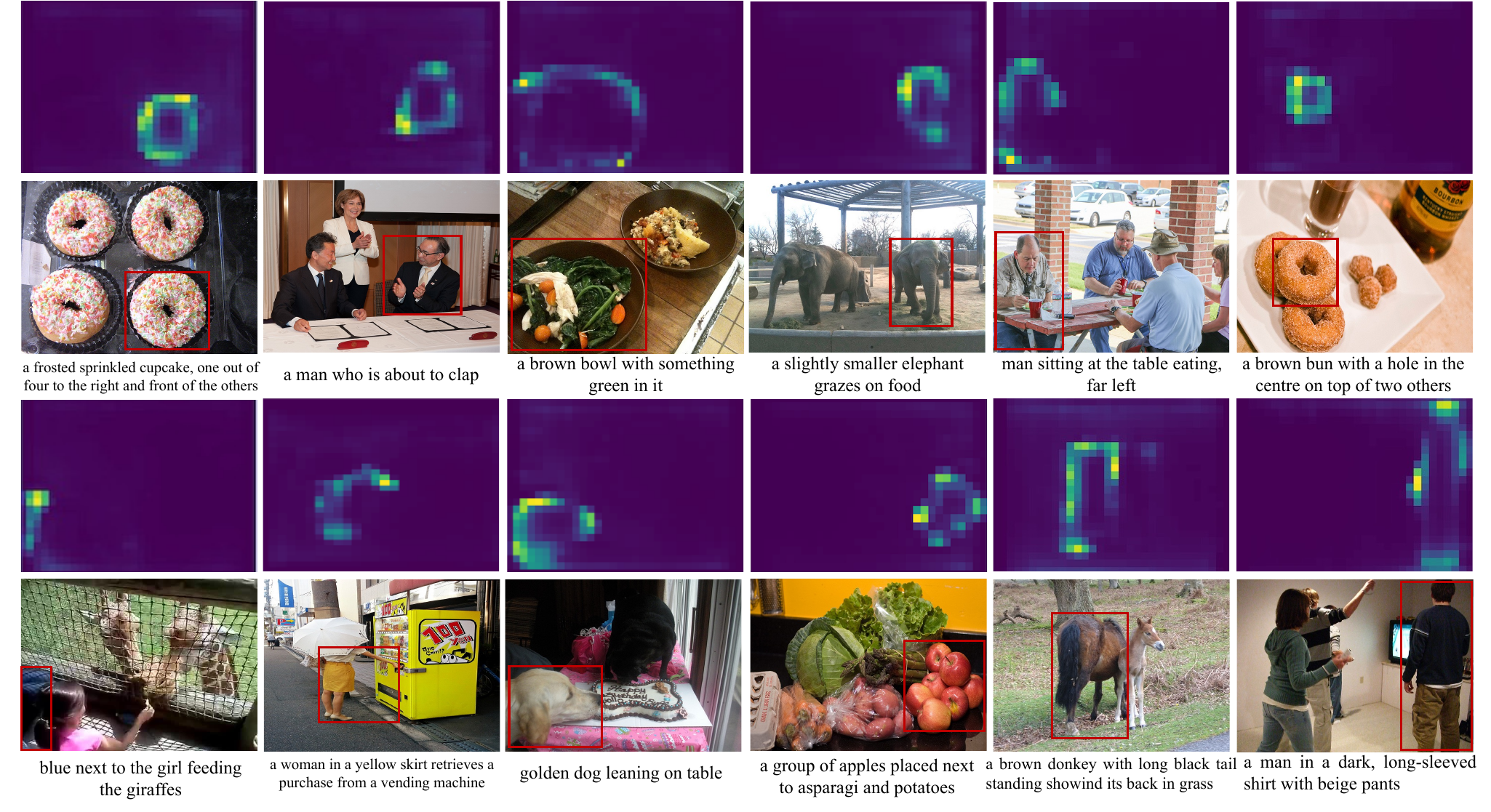}
  \caption{Visualization of attention maps.
  The \textcolor{red}{red} boxes represent the prediction of our framework.
  }
  \label{fig:reg}
\end{figure*}

\subsection{Ablation Analysis}
In this section, we conduct a series of ablation studies to further investigate the relative importance and specific contributions of each component in our proposed framework.
We use ResNet-50 \cite{resnet} and six-layer BERT \cite{bert} as visual and linguistic encoders, respectively, to conduct ablation experiments on the val set of RefCOCOg-google.

\subsubsection{The effect of the number of visual and linguistic features}
We study the effect of the number of visual and linguistic features without self-paced sample informativeness learning, as illustrated in \cref{tab:number_features}.
To avoid the influence of the number of CA$^{2}$M, we uniformly set a total of six and three CA$^{2}$Ms in the grouping and merging cross-modal interactions, respectively.
Specifically, we set the baseline that only uses the last layer of visual and linguistic features, and gradually increases the number of visual and linguistic features.
As shown in \cref{tab:number_features}, it can be observed that with the increase of multi-grained visual and linguistic features, the performance of the model improves significantly, which indicates the importance of multi-grained visual and linguistic information for REC.

\subsubsection{The effect of the number of CA$^{2}$M}
We study the effect of the number of CA$^{2}$M without self-paced sample informativeness learning, as illustrated in \cref{tab:number_layers}.
To avoid the influence of the number of visual and linguistic features, we uniformly use three layers of visual and linguistic features.
Specifically, we set the baseline that uses a total of three and one CA$^{2}$Ms in the grouping and merging cross-modal interactions, respectively.
As shown in \cref{tab:number_layers}, as the number of layers increases, the performance of the network improves.
In addition, the experimental results in the fifth and sixth rows of \cref{tab:number_layers} demonstrate that grouping and merging cross-modal interactions can help the network progressively understand multi-grained cross-modal information and achieve higher accuracy.

\subsubsection{The effect of $\lambda$ and $\sigma$}
We study the effect of $\lambda$ and $\sigma$ in self-paced sample informativeness learning, as shown in \cref{fig:lambda_sigma}.
To avoid the influence of the number of features and CA$^{2}$M, we uniformly use three layers of visual and linguistic features, and set a total of six and three CA$^{2}$Ms in the grouping and merging cross-modal interactions, respectively.
During the experiment, we fixed $\lambda$ or $\sigma$, and perform the experiment on the other parameter.
As shown in \cref{fig:lambda_sigma}, we can get the best experimental results when $\lambda$ and $\sigma$ are set to 0.75 and 2.5, respectively.
With the help of self-paced sample informativeness learning, the overall performance of the model increases from 72.86\% to 73.87\%.
In addition, on the basis of the above parameters, we plot the variation of the $e^{\frac{(\mathcal{L}-\mu)^{2}}{\sigma^{2}}}$ during training, as shown in \cref{fig:epoch_sigma}.

\subsubsection{Comparison of CA$^{2}$M and Transfomer Encoder Layer}
We perform a comparison of CA$^{2}$M and transformer encoder layer \cite{transformer}.
Concretely, we uniformly use three layers of visual and linguistic features, equip self-paced sample informative learning, and set a total of six and three cross-modal interaction blocks in the grouping and merging cross-modal interactions, respectively.
We conduct experiments using CA$^{2}$M or transformer encoder layer as cross-modal interaction block.
For the transformer encoder layer, same as state-of-the-art methods \cite{transvg,mdetr,referring-transformer,pfos}, we concatenate visual and linguistic features, and perform self-attention cross-modal interaction.
\cref{tab:cmp} shows the comparison of CA$^{2}$M and transformer encoder layer on RefCOCO, RefCOCO+, RefCOCOg and ReferItGame, which validate the effectiveness of our proposed CA$^{2}$M.
Because of the cross-attention, CA$^{2}$M has a faster inference speed.
Concretely, given linguistic sequence length $l$, visual sequence length $hw$, and common feature space dimension $c$, the cross-attention computational complexity of CA$^{2}$M is $\mathcal{O}(hwlc)$, while the self-attention computational complexity of transformer encoder layer is $\mathcal{O}((hw+l)^2c)$. 
Considering $hw>>l$, although CA$^{2}$M performs cross-attention twice, it is still faster than the self-attention of the transformer encoder layer. 
And due to the structure of two cross-modal interactions, the parameters of CA$^{2}$M are larger.

\subsection{Qualitative Results}
To better demonstrate the effectiveness of our proposed method, we show the visualization results of our framework and the baseline on the RefCOCOg-google val set in \cref{fig:visual}.
Specifically, our proposed framework utilizes multi-grained cross-modal attention and self-paced sample informativeness learning while the baseline only uses features extracted from the last layer of visual and linguistic encoders.
As shown in \cref{fig:visual}, we can observe that our framework can holistically understand both visual and linguistic modalities for accurate target localization.
We also visualize the [CLS] token's attention to visual tokens, as shown in \cref{fig:reg}.
[CLS] tokens incorporate the multi-grained information of visual and linguistic modalities, and accurately locate the visual regions corresponding to the referring expressions.
\section{Conclusion}
\label{sec5}
In this paper, we propose a Self-paced Multi-grained Cross-modal Interaction Modeling framework, which improves the language-to-vision localization ability by aggregating multi-grained information from different modalities and extracting abundant knowledge from hard examples.
To be specific, we construct a transformer-based multi-grained cross-modal attention, which effectively utilizes the inherent multi-grained information in visual and linguistic encoders.
In addition, we design a self-paced sample informativeness learning to adaptively realize network learning for samples with abundant multi-grained information.
Extensive experiments have demonstrated the effectiveness of our framework.
In the future, we will explore how to further exploit the multi-grained information of visual and linguistic modalities to improve the reasoning ability of the network.

\section*{Acknowledgment}
This work is supported in part by the National Key Research and Development Program of China under Grant 2020AAA0107400, National Natural Science Foundation of China under Grant U20A20222, National Science Foundation for Distinguished Young Scholars under Grant 62225605, Zhejiang Key Research and Development Program under Grant 2023C03196, the Zhejiang Provincial Natural Science Foundation of China under Grant LR19F020004, and sponsored by CCF-AFSG Research Fund, CCF-Zhipu AI Large Model Fund (CCF-Zhipu202302) as well as The Ng Teng Fong Charitable Foundation in the form of ZJU-SUTD IDEA Grant, 188170-11102.

\bibliographystyle{IEEEtran}
\bibliography{IEEEabrv,bibli}

\begin{IEEEbiography}[{\includegraphics[width=1in,height=1.25in,clip,keepaspectratio]{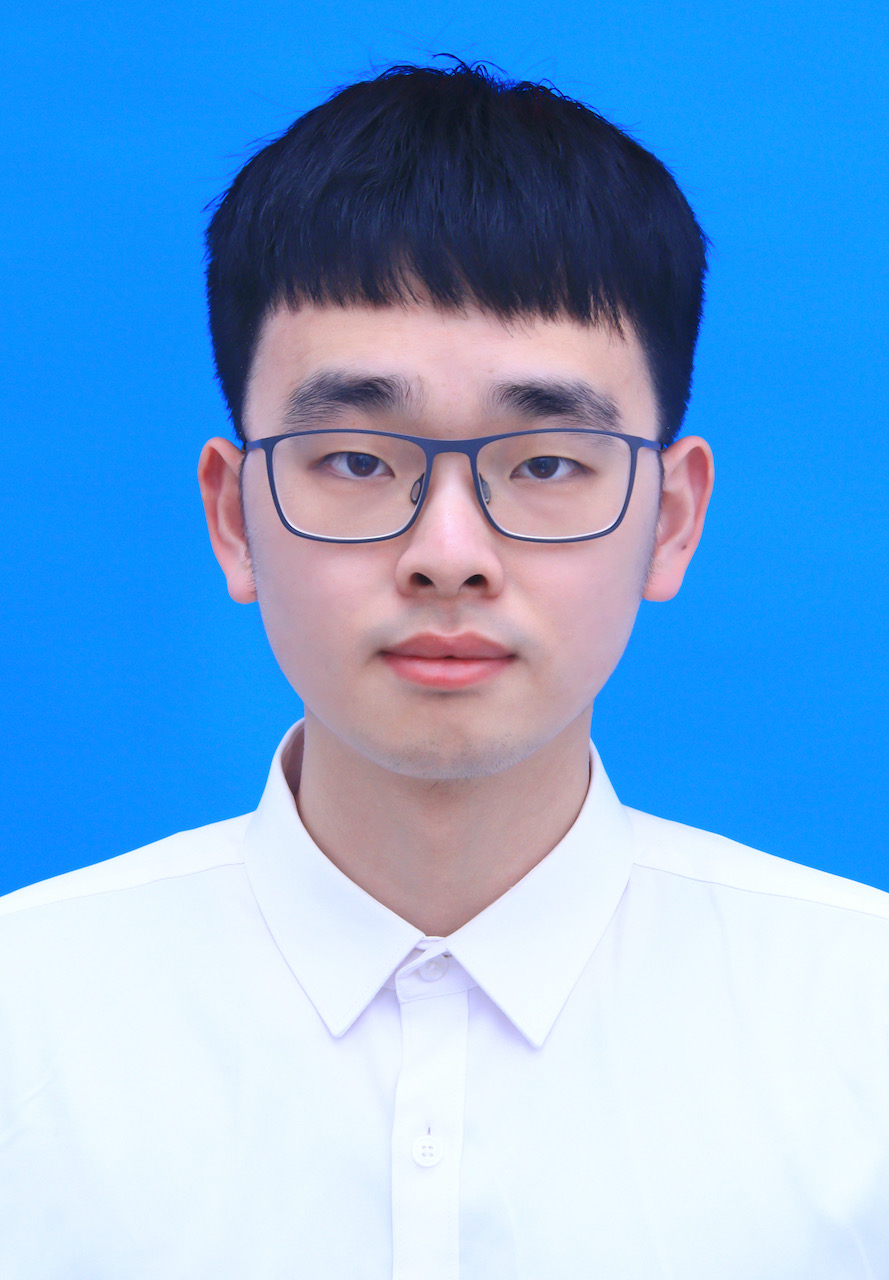}}]{Peihan Miao}
  received the B.S. degree in computer science and technology from Zhejiang Gongshang University, China, in 2021. He is currently pursuing the M.S. degree with the College of Software Technology, Zhejiang University, China, under the supervision of Prof. X. Li. His current research interests include computer vision, multi-modal learning and referring expression comprehension.
\end{IEEEbiography}

\begin{IEEEbiography}[{\includegraphics[width=1in,height=1.25in,clip,keepaspectratio]{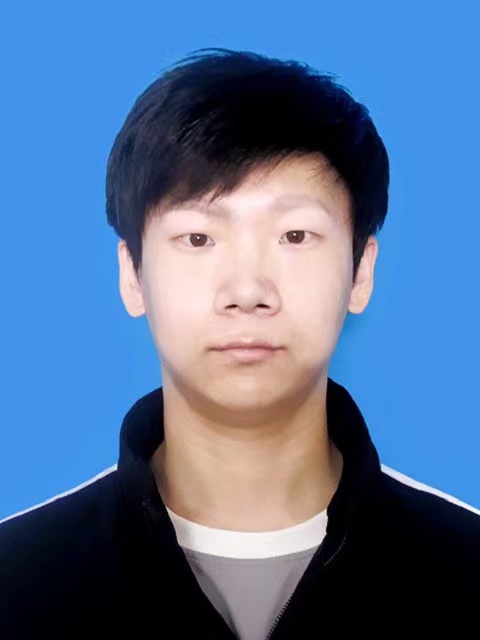}}]{Wei Su}
  received the B.S. degree in electronic science and technology and the M.S. degree in computer science and technology from Northwestern Polytechnical University, Xi’an, China, in 2018 and 2021, respectively. He is currently pursuing the Ph.D. degree at Zhejiang University, China, under the supervision of Prof. X. Li. His current research interests include visual grounding, dynamic neural networks, and visual language models.
\end{IEEEbiography}

\begin{IEEEbiography}[{\includegraphics[width=1in,height=1.25in,clip,keepaspectratio]{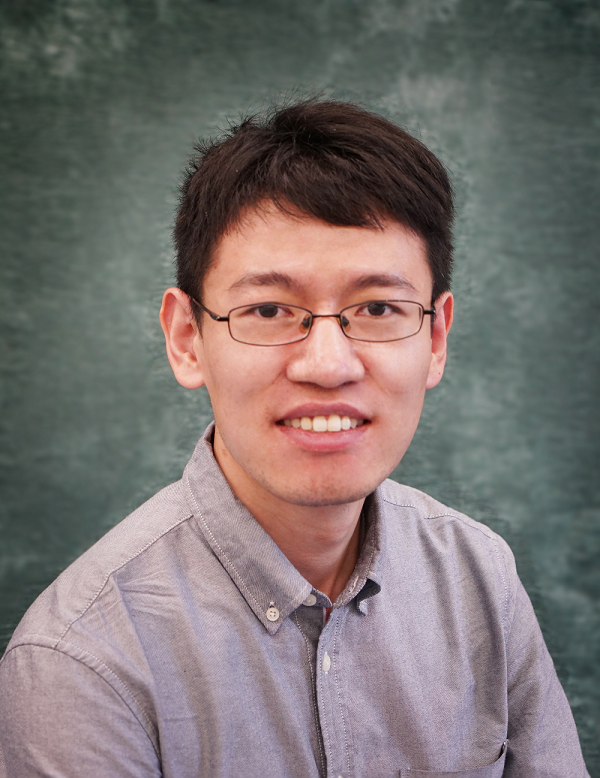}}]{Gaoang Wang}
  received the B.S. degree at Fudan University in 2013, the M.S. degree at the University of Wisconsin-Madison in 2015, and the Ph.D. degree from the Information Processing Laboratory of the Electrical and Computer Engineering department at the University of Washington in 2019. He joined the international campus of Zhejiang University as an Assistant Professor in September 2020. He is also an Adjunct Assistant Professor at UIUC. His research interests are computer vision, machine learning, artificial intelligence, including multi-modal learning, multi-object tracking, representation learning, and active learning.
\end{IEEEbiography}

\begin{IEEEbiography}[{\includegraphics[width=1in,height=1.25in,clip,keepaspectratio]{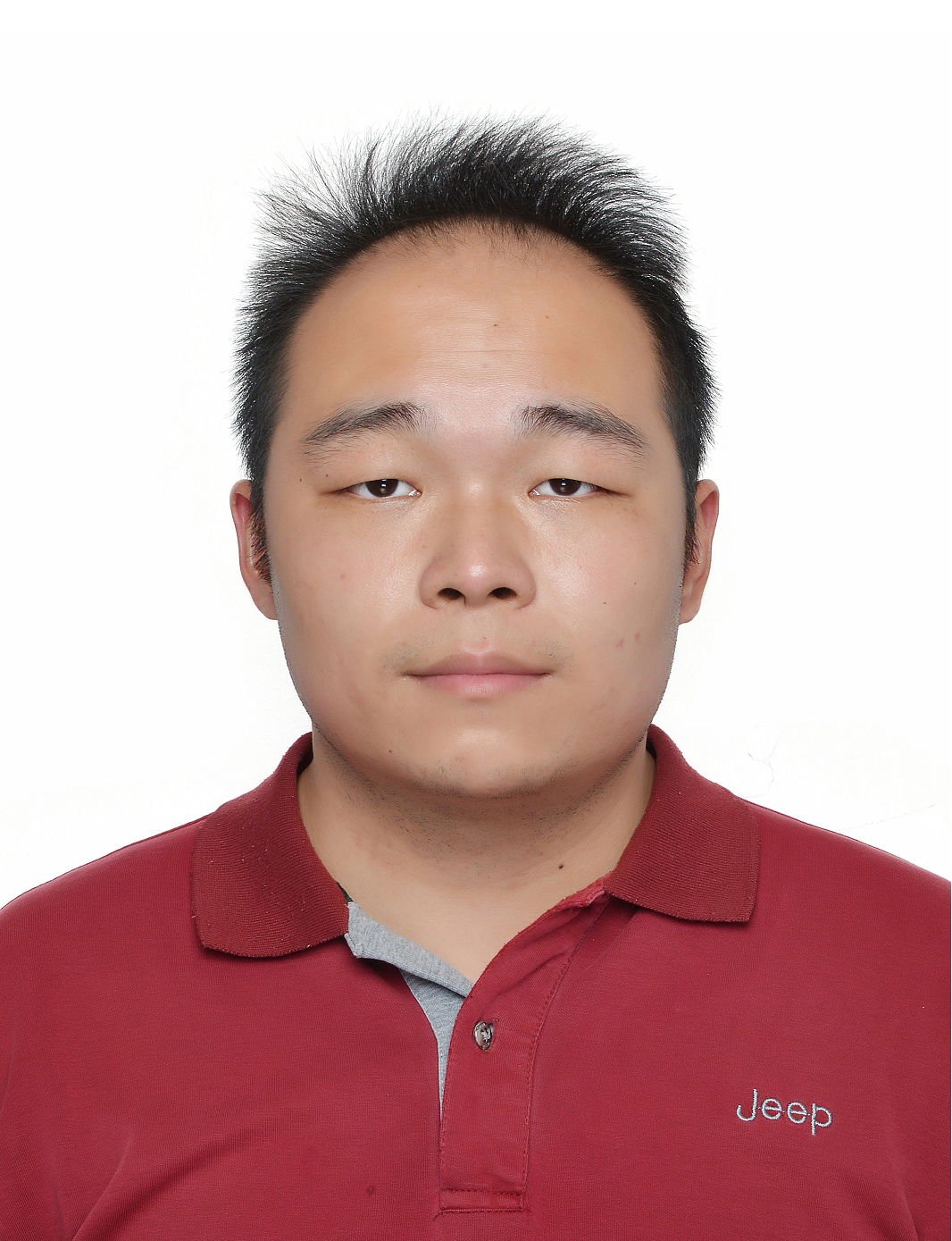}}]{Xuewei Li}
  received his bachelor of engineering in 2019 from Zhejiang University, China. He is currently a Ph.D. candidate at Zhejiang University. His current research interests include knowledge distillation, scene graph generation, panoramic semantic segmentation, referring expression comprehension, and diffusion models.
\end{IEEEbiography}

\begin{IEEEbiography}[{\includegraphics[width=1in,height=1.25in,clip,keepaspectratio]{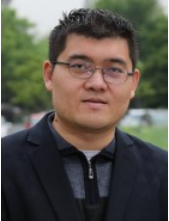}}]{Xi Li}
  received the Ph.D. degree from the National Laboratory of Pattern Recognition, Chinese Academy of Sciences, Beijing, China, in 2009. From 2009 to 2010, he was a Post-Doctoral Researcher with CNRS Telecom ParisTech, France. He was a Senior Researcher with the University of Adelaide, Australia. He is currently a Full Professor with Zhejiang University, China. His research interests include visual tracking, compact learning, motion analysis, face recognition, data mining, and image retrieval.
\end{IEEEbiography}

%









\end{document}